\documentclass{article}
\usepackage{adjustbox}

\usepackage{PRIMEarxiv}
\usepackage{appendix}

\usepackage[utf8]{inputenc} 
\usepackage[T1]{fontenc}    
\usepackage{hyperref}       
\usepackage{url}            
\usepackage{booktabs}       
\usepackage{amsfonts}       
\usepackage{nicefrac}       
\usepackage{microtype}      
\usepackage{lipsum}
\usepackage{fancyhdr}       
\usepackage{graphicx}       
\usepackage{subcaption}
\usepackage{placeins}
\usepackage{adjustbox}  
\usepackage{xcolor,soul}
\usepackage{amsmath}
\sethlcolor{yellow!30}
\graphicspath{{media/}}     

\title{Privacy-Preserving Synthetic Review Generation with Diverse Writing Styles Using LLMs

}

\author{
  Tevin Atwal, Chan Nam Tieu, Yefeng Yuan, Zhan Shi, Yuhong Liu\\
  Department of Computer Science and Engineering \\
  Santa Clara University \\
  \texttt{\{tatwal, ctieu, yyuan4, ashi2, yhliu\}@scu.edu} \\
   \And
  Liang Cheng \\
  eBay \\
  \texttt{liacheng@ebay.com} \\
}

\begin{document}
\maketitle

\begin{abstract}
The increasing use of synthetic data generated by Large Language Models (LLMs) presents both opportunities and challenges in data-driven applications. While synthetic data provides a cost-effective, scalable alternative to real-world data to facilitate model training, its diversity and privacy risks remain underexplored. Focusing on text-based synthetic data, we propose a comprehensive set of metrics to quantitatively assess the diversity (i.e., linguistic expression, sentiment, and user perspective), and privacy (i.e., re-identification risk and stylistic outliers) of synthetic datasets generated by several state-of-the-art LLMs. Experiment results reveal significant limitations in LLMs' capabilities in generating diverse and privacy-preserving synthetic data. Guided by the evaluation results, a prompt-based approach is proposed to enhance the diversity of synthetic reviews while preserving reviewer privacy.

\end{abstract}

\keywords{Synthetic Data Generation \and Writing Style Diversity \and Privacy \and Prompt Optimization a\and LLM}

\section{Introduction} \vspace{-2mm}
The recent advancements in AI technologies have significantly increased the demand for vast amounts of data for training. In contexts where acquiring high-quality, diverse, and privacy-preserving datasets can be prohibitively expensive or legally constrained, synthetic data offers a promising alternative by reducing the dependency on real-world data while addressing privacy concerns \cite{guo2024generative}. Despite its advantages, synthetic data often lacks the linguistic diversity and variability that real data offers, potentially leading to biased outcomes and misrepresentation in AI systems \cite{hao2024synthetic}. Moreover, recent studies show that LLM-generated synthetic data may unintentionally include original data records from the training stage, sparking widespread concerns about LLMs' ability to memorize and regurgitate sensitive training data \cite{carlini2021extracting, das2025security}. In light of these issues, enhancing the diversity and privacy of synthetic data has become essential for ensuring ethical AI development.

As a first step to tackle these issues, this study aims to quantitatively assess the diversity and privacy of LLM-generated synthetic data, explore the trade-off, and propose potential solutions to enhance diversity while preserving user privacy. In particular, we focus on online product review data since they are particularly rich in users' customized information and variability due to their mix of sentiment, linguistic expression, and user perspective, making them an ideal context for this study. 

A comprehensive set of metrics is proposed to evaluate (1) diversity from lexical, semantic, and sentiment aspects, and (2) privacy risks from the presence of personal identifiable content and stylistic uniqueness at the user level. By applying the proposed metrics on both real user data and several synthetic data sets generated by different LLMs, we discuss the results and key findings in detail. Based on the evaluation feedback, we introduce an automatic prompt optimization pipeline that adaptively updates generation instructions according to failed metric evaluations. Specifically, this pipeline stacks targeted prompt refinements for each diversity or privacy criterion (e.g., lexical variation or outlier control), resulting in higher-quality, more balanced synthetic review datasets. Compared to static prompting, this dynamic strategy improves overall coverage across desired attributes with minimal manual tuning.

This study contributes to the broader discourse on the ethical and practical considerations surrounding synthetic data, particularly in review data applications, aiming to foster more inclusive and safe AI systems. 
The major contributions of this work are summarized as follows:
\begin{itemize}
\vspace{-2mm}
    \item We propose a comprehensive set of metrics to quantitatively evaluate the diversity and privacy of user review data. The effectiveness of these metrics are demonstrated using an Amazon Review dataset containing over 2.5 million real user reviews. \vspace{-1mm}
    \item We generate a diverse set of synthetic review data through multiple state-of-the-art LLMs, including GPT‑4o and Claude 3.7 Sonnet, and evaluate the diversity and privacy of these synthetic data based on the proposed metrics. The evaluation results are analyzed and compared to investigate different model's capabilities in synthetic data generation, and tradeoffs between writing style diversity and privacy risks of author re-identification. \vspace{-1mm}
    \item An iterative prompt-based enhancement approach is proposed to enhance writing style diversity while preserving reviewer privacy. The proposed method features an automatic prompt optimization component that dynamically adapts instructions based on metric failures, resulting in more semantically rich, stylistically varied, and privacy-compliant reviews. \vspace{-1mm}
\end{itemize}

\section{Related Work}
\vspace{-2mm}

LLMs are vital for generating synthetic text, enhancing or substituting real data under privacy constraints. However, substantial challenges persist, particularly concerning the complexity, authenticity, and representational diversity inherent to genuine textual data \cite{hao2024synthetic}.
For example, Hämäläinen et al. \cite{hämäläinen2023evaluating} revealed that synthetic data exhibited notably lower diversity and included specific, identifiable anomalies. In addition, Carlini et al. \cite{carlini2021extracting} highlight cases where LLMs memorize and reproduce exact phrases or PII from training data, exposing a clear vulnerability to targeted extraction of sensitive information. These issues have motivated researchers to establish rigorous metrics and evaluation protocols to assess the quality of synthetic data. 

 Researchers frequently quantify lexical and structural diversity using metrics such as Distinct-$n$, Self-BLEU, entropy, sentence lengths, and semantic diversity through embedding-based clustering approaches \cite{chen2024}.  Shaib et al. \cite{Shaib2024} established a standardized framework for diversity measurement by comparing metrics based on the average similarity between word pairs and various token/type ratios. In our work, we have adopted this structured approach for the evaluation of lexical diversity. Moreover, we propose multiple semantic diversity metrics based on minimum spanning tree (MST) to capture the semantic diversity aspect.  Yuan et al. \cite{yuan2019} assessed the validity of synthetic review text by correlating the distribution of review sentiment with the corresponding rating values, inspiring the design of sentiment diversity metrics in this study. Montahaei et al. \cite{Montahaei2019} concentrated on the joint evaluation of quality and diversity in text generation models. Lautrup et al. \cite{Lautrup_2024} jointly evaluated the fidelity and privacy-preserving integrity of synthetic data. Extending this line of inquiry, we jointly evaluate model diversity and privacy to further elucidate the trade-offs between these two aspects, thereby facilitating a more profound understanding of their interplay.

To enhance the quality of synthetic data generation, several studies have integrated iterative refinement techniques, nuanced prompt engineering, and explicit bias control mechanisms \cite{app12094619}. Barbierato et al. \cite{app12094619} introduced a probabilistic network-based methodology that explicitly manages biases and fairness within synthetic datasets, allowing for precise manipulation of feature relevance and interdependencies.  
Whitney and Norman \cite{Whitney_2024} caution against potential risks such as diversity-washing, wherein synthetic datasets may superficially appear diverse but fail to address underlying representational biases and issues of consent circumvention. To mitigate these concerns, we propose an iterative prompt-based approach grounded in our well-defined comprehensive metric evaluation, encompassing both diversity and privacy considerations. This approach demonstrates favorable performance while offering enhanced flexibility.

\begin{figure}[t]
\centering
\includegraphics[width=0.8\linewidth]{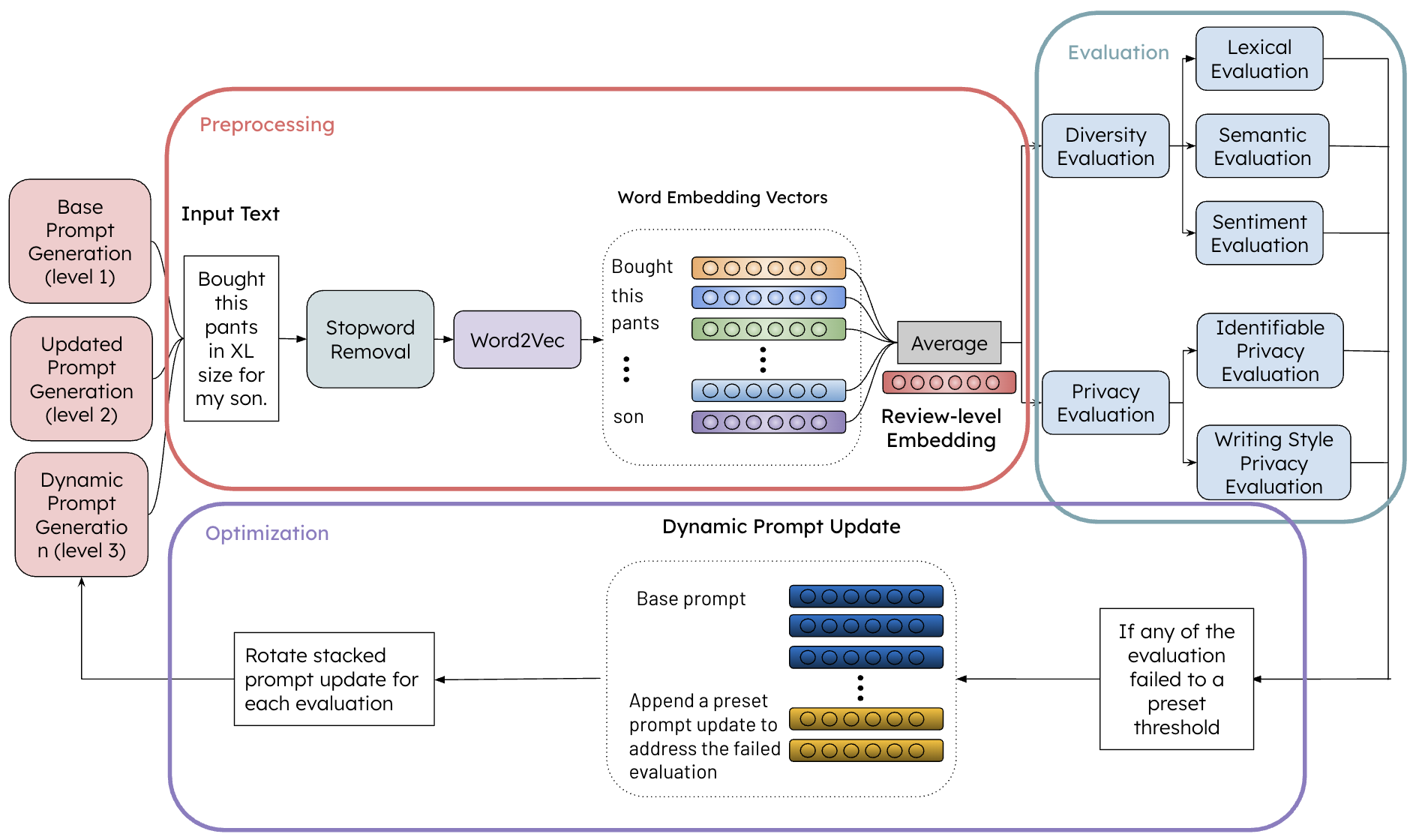}
\caption{Overview of the Evaluation Framework}
\label{fig:preprocessing_workflow}
\end{figure}
\vspace{-1mm}

\section{Approach}
\label{sec:headings}

In this section, we discuss the proposed evaluation framework, as shown in Figure \ref{fig:preprocessing_workflow}, which includes some preprocessing steps for review embeddings, a comprehensive set of quantitative metrics proposed from diversity and privacy perspectives, and a dynamic prompt generation module to adjust prompt contents based on  

\subsection{Preprocessing Workflow}\label{sec:embedding}
To quantitatively assess the diversity and privacy of synthetically generated review data, this research implements a structured preprocessing and vectorization workflow, as shown in Figure \ref{fig:preprocessing_workflow}. The workflow begins with removing stopwords such as "the", "is", and "at", which frequently appear in text but provide little meaningful semantic information. Reviews are then converted into numerical representations using the Word2Vec embedding model. Instead of using generalized pre-trained embedding models, which might introduce irrelevant external semantic biases, training Word2Vec specifically on the dataset under study ensures the embeddings precisely capture the linguistic characteristics specific to the dataset. To represent each review quantitatively, a sentence-level embedding is computed by averaging all word vectors within each review. Finally, the embeddings are fed into the diversity and privacy evaluation modules for further analysis. Please refer to Appendix A for more details.

\subsection{Evaluation Metrics}
To systematically evaluate diversity and privacy risks of the dataset, we propose a comprehensive set of metrics. Specifically, diversity is evaluated from three aspects, (1) including word-level lexical richness, (2) review-level semantic diversity, and (3) sentiment diversity. Two complementary privacy metrics (1) the presence of personal identifiable content, and (2) stylistic uniqueness at the user level, are proposed to measure two types of text privacy risks: \textit{unintentional memorability} and \textit{authorship attribution}~\cite{SHAHRIAR2025104358}. Both metrics quantify signals that can expose individual users to privacy vulnerabilities when their data is included in training sets for generative models.

\subsubsection{Word-level Lexical Richness}
We propose to evaluate word-level lexical richness by utilizing an N-gram-based approach, which analyzes sequences of contiguous words ranging from unigrams (single words) through pentagrams (five-word sequences). This multilevel approach provides critical insights into both local lexical diversity and broader linguistic patterns. Specifically, we adopted two metrics for each N-gram level: the \textbf{lexical uniqueness ratio} (i.e., marked as $L_{r}$), which is the proportion of unique N-grams relative to the total number of N-grams identified (i.e., representing lexical richness), and \textbf{normalized lexical entropy} (i.e., marked as $H_n$), which captures the distributional complexity of these lexical elements (i.e., representing structural variability). A high lexical uniqueness ratio coupled with low normalized lexical entropy might indicate lexical diversity, but uneven distribution of phrase structures. Whereas high scores in both metrics indicate both lexical and structural diversity.

\subsubsection{Review-level Semantic Diversity}
In this study, we propose to evaluate the semantic diversity among reviews in the embedding space. Specifically, through the preprocessing shown in Figure \ref{fig:preprocessing_workflow}, we retrieve the embedding of each review, which serves as the basis for the two evaluation metrics below. 

\textbf{Semantic Ratio}: the ratio of distinct embedding vectors to the total number of reviews. This metric evaluates the degree of repetition in the dataset. A higher ratio indicates fewer repetitions and greater semantic uniqueness among reviews, while a lower ratio implies substantial repetition, suggesting limited diversity.

\begin{figure}[t]
    \centering
\includegraphics[width=0.6\linewidth]{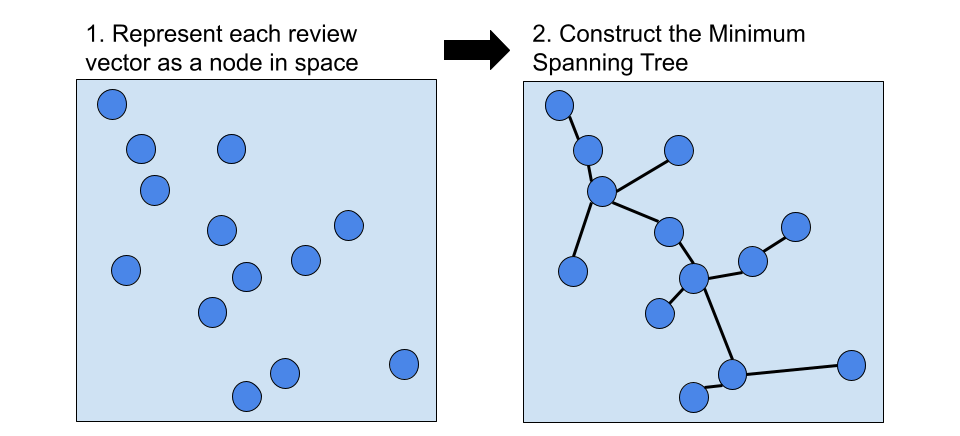}
    \caption{Visual representation of the Minimum Spanning Tree where each node represents a single review vector and the edges are the calculated cosine distance between review vectors.}
    \label{fig:MST_graphic}
\end{figure}

\textbf{Average MST Edge Length}. Inspired by \cite{cox2021directed}, we adopt a minimum spanning tree (MST)-based approach to evaluate the review-level semantic diversity. The MST is established based on pairwise semantic distances (i.e., computed as cosine distance) between review embeddings, as shown in Figure \ref{fig:MST_graphic}. MST is adopted since it connects all reviews in a semantic network with the minimal number of edges while maintaining connectivity, ensuring diverse reviews remain linked without redundant connections. It also captures the structural relationships between reviews in a way that highlights the structural variability. The average MST edge length provides insight into the average semantic distance between connected reviews. Higher average edge lengths suggest that reviews are less similar, thus reflecting increased semantic richness and reduced redundancy.
\vspace{1mm}

\vspace{-3mm}

\subsubsection{Sentiment Diversity}

In this study, we consider sentiment diversity as an important metric to measure whether synthetic reviews can realistically represent a variety of human-like patterns of satisfaction or dissatisfaction across different rating levels. The rationale is that synthetic reviews should be generated with diverse sentiments, so that reviews with 1-star ratings have a higher percentage of negative sentiment, while reviews with 5-star ratings have a lower percentage of negative sentiment. 

The computation process is as follows. First, we group all synthetic reviews into five segments based on their corresponding rating values. Within each segment, a sentiment analysis is performed on each review. Specifically, the Flair sentiment classifier \cite{akbik2019flair}, which leverages distilBERT embeddings, is chosen due to its robustness and precision in classifying nuanced textual sentiments into binary values (i.e., positive or negative). Next, we compute the sentiment distribution within each segment and evaluate whether the sentiment distribution across segments can distinctively reflect user satisfaction. Specifically, within each segment, we compute the sentiment score as the percentage of reviews with positive sentiment. 

To evaluate the sentiment diversity of synthetic reviews, we established an ideal linear sentiment distribution for benchmarking. This benchmark reflects an intuitive expectation that sentiment scores should monotonically increase with ratings (e.g., rating 1 predominantly negative, rating 5 predominantly positive). 
Given a synthetic review data set, comparing its sentiment distribution across segments to this ideal linear distribution directly reveals how well these synthetic reviews align with the ideal human sentiment patterns. Larger deviations highlight the bias in the sentiment of synthetic reviews. Therefore, we propose to measure sentiment diversity as an average of Mean Absolute Error (MAE) across different rating values, as shown in Equation (\ref{eq:MAE_sum}).

\vspace{-5mm}
\begin{equation}
D_{sen} = \frac{1}{5}\sum_{i=1}^5 (1-|y_i-\bar{y}_i|) \label{eq:MAE_sum}
\end{equation}
where $y_i$ represents the percentage of positive reviews for segment $i$ (i.e., with rating value equal to $i$), and $\bar{y}_i$ represents the sentiment scores of the linear benchmark.

\subsubsection{Contextual Identifiers in Text: Named Entities and Nominal Mentions}
This metric aims to quantify the presence of personally identifiable and contextually revealing information within user-generated reviews. We focus on two types of linguistic features that contribute to privacy risk: named entities, which often reflect specific real-world references (e.g., “Jeff Bezos”, “Seattle”, “Christmas”), and nominal mentions, which signal the presence of personal narratives or roles (e.g., family members, recipients, subjects of sentences). These elements serve as proxies for evaluating the risk of unintentional memorability in LLMs.

For both named entities and nominal mentions, we evaluate two complementary signals: total count and density. The total count captures reviews that are long and context-rich—where multiple sensitive terms may accumulate and be leveraged for re-identification—while density highlights shorter, more fragmented reviews that contain a high concentration of revealing terms, which may still enable associative inference, even from a single word. Based on named entities and nominal mentions, we can evaluate potential privacy exposure in the dataset at the content-level. High frequencies and densities of these elements may indicate the presence of personally identifiable or disclosing information that generative models could memorize during training. As such, these metrics serve as indicators of potential unintentional memorability, especially in LLMs that have been trained on user-generated content. They also provide a basis for the comparison of privacy properties between original and synthetic datasets.

\begin{table}[t]
\centering
\caption{Representative examples of privacy-relevant reviews across high entity/nominal count and density scenarios. Named entities are underlined; nominal mentions are highlighted.}
\label{tab:sensitive_examples}
\begin{tabular}{p{0.68\linewidth} p{0.08\linewidth} p{0.08\linewidth}}
\toprule
\textbf{Review Text} & \textbf{Entity Density} & \textbf{Nominal Density} \\
\midrule
“\hl{I} love this baseball \hl{cap}.  \hl{I} graduated from \underline{the \hl{University} of \hl{Hawaii}} with \hl{my} \underline{\hl{Bachelor}}'s \hl{degree}...and \hl{I} love advertising \underline{\hl{Hawaii}} on the \hl{top} of \hl{my} \hl{head}!  \underline{The many years} \hl{I} lived in \underline{\hl{Hawaii}} ~ \hl{it} was/is absolutely gorgeous, calm, safe, friendly and multi-ethnic.  Great memories...thus, a happy cap to bring back happy \hl{memories}.” & 0.086 & 0.293 \\\hline \hline
“\hl{I} have plantar \hl{fasciitis} and have been trying and using various compression \hl{socks} and sleeves... \hl{I} ordered the large/extra-large because \hl{I} take a \underline{9 to 9.5} \hl{shoe}... \hl{I}'m passing \hl{them} off to my \hl{boyfriend}... \hl{These} are not only ‘fun’ but \hl{they} are medically helpful for \hl{my} plantar \hl{fasciitis}...” & 0.025 & 0.284 \\\hline \hline
“Bought \hl{this} in \underline{\hl{XL}} for \hl{my} \hl{11yo} \hl{who} is \underline{5'8} and \underline{110}.” & 0.250 & 0.417 \\ \hline \hline
“\hl{My} \hl{granddaughter} loves \hl{these}!” & 0.0 & 0.750 \\
\bottomrule
\end{tabular}
\end{table}

\subsubsection{Stylistic Outlier Detection via Embedding-based User Profiling}
This metric focuses on capturing stylistic uniqueness at the user level to evaluate the risk of \textit{authorship attribution}. Users with highly distinctive writing styles may be more vulnerable to re-identification attacks, particularly when generative models preserve these stylistic cues in synthetic outputs.

Specifically, we propose to capture users' writing styles by constructing user-level embedding vectors, which is computed as the mean of all the review embeddings of a user. A user is considered a stylistic outlier if his/her embedding is both \textit{globally} and \textit{locally} distant from others. A user is globally distant if his/her average similarity to the other users is significantly lower than that of the majority users, and is locally distant if his/her distance from the nearest neighbor is far. For a given dataset, we propose to use the number of identified stylistic outliers (i.e., $|| \mathcal U ||$) to quantitatively represent the privacy risks, where more users with distinctive writing styles indicate higher privacy risks.

\paragraph{Global Similarity Analysis}
To assess the global stylistic distance of a user from others, we compute the average cosine similarity of a user based on his/her similarity to all other users in the dataset (i.e., marked as $S_i$). Then, the z-score (i.e., $Z_i$) as each user's similarity score with respect to the distribution of all users' average similarities is computed as:

\vspace{-5mm}
\begin{equation}
Z_i = \frac{S_i - \mu_S}{\sigma_S}
\end{equation}
where \( \mu_S \) and \( \sigma_S \) are the mean and standard deviation of the \( S_i \) values across all users. A low z-score indicates that the user is globally dissimilar from the majority, and is therefore a candidate outlier for further local inspection. Specifically, we define a global threshold $\theta_{g}$ and extract all users with their z-scores below this threshold as global distinct users (i.e., $\mathcal U_g$) for additional inspection. 

\vspace{-5mm}
\begin{equation}
    \mathcal U_g =\{u_i |Z_i \le \theta_{g}\} \label{eq:global_dis} 
\end{equation}

\paragraph{Local Distance Filtering}
Our local distance-based filtering step is motivated by the principle of \(k\)-anonymity. A user should only be considered a true stylistic outlier—and therefore at heightened risk of authorship attribution—if their writing style does not resemble that of any other user in the dataset, effectively corresponding to \(k = 1\). Relying solely on global similarity scores can mistakenly classify side clusters of users (numbering in the hundreds or even thousands) as outliers, simply because they are distant from the general population, even though they form a coherent group with one another. Since our goal is to detect true outliers rather than small but internally consistent subgroups, we refine our candidate set by inspecting each user's distance to their nearest neighbor.

To implement this step, we apply a local distance-based filter to the set of users with globally low z-scores (i.e., $\mathcal U_g$). For each selected user $i$ from $\mathcal U_g$, we compute its cosine distance to its nearest neighbor in the embedding space. The nearest neighbor distance $d^{nn}_i$ is computed as:

\vspace{-5mm}
\begin{equation}
d^{nn}_i = \min_{j \ne i} d_{cos}(u_i, u_j)
\end{equation}

Note that user $j$, as the most similar user to user $i$, can be any user in the entire dataset, not necessarily from $\mathcal U_g$. After identifying the nearest neighbors for all globally distant users, we rank them by their $d^{nn}_i$ values in descending order. User $i$ will be finalized as a stylistic outlier if: 

\vspace{-5mm}
\begin{equation}
\mathcal U_o = \{u_i \in \mathcal U_g |d^{nn}_i \ge \theta_l \}
\end{equation}

where $\theta_l$ is the local threshold.

The users identified by this outlier detection method exhibit distinctive or idiosyncratic writing styles compared to the rest of the dataset. By isolating these individuals, we can assess which portions of the data are most likely to exert a disproportionate influence on authorship attribution risk. This analysis provides a useful view for characterizing the stylistic privacy profile of the dataset and can inform data filtering measures, user anonymization, or targeted privacy preservation strategies.

\subsection{Adaptive Prompt Optimization via Evaluation-Guided Feedback}\label{sec:ada_prompt}

To improve the quality and diversity of synthetic product reviews generated by LLMs, we introduce a dynamic prompt optimization pipeline that adapts its instructions based on multi-metric feedback. Rather than relying on a fixed prompt applied across all data generation cycles, our method incrementally adjusts the prompt in response to failures in predefined evaluation metrics. This approach enables iterative refinement of outputs while avoiding overfitting to a static instruction template.

The system begins with a base prompt that instructs the LLM to generate realistic and varied product reviews. After each generation cycle, the resulting batch is evaluated along six core dimensions:

\begin{itemize}
    \item \textbf{Word-level lexical diversity:} Ensuring varied word usage and expressions across reviews.
    \item \textbf{Review-level semantic diversity:} Encouraging unique contexts, use cases, and user experiences.
    \item \textbf{Sentiment diversity:} Matching the emotional tone to the review rating.
    \item \textbf{Outlier detection:} Identifying semantically or stylistically anomalous reviews.
    \item \textbf{Uniqueness:} Avoiding duplication of previously generated content.
    \item \textbf{Length distribution:} Enforcing a fixed distribution of review lengths, such as 25\% very short (1--10 words), 40\% medium (11--40 words), 25\% long (41--80 words), and 10\% extra-long (80+ words)
\end{itemize}

Each evaluation metric operates independently and informs the prompt adjustment mechanism. If a generated batch fails any metric, a corresponding instruction is drawn from a predefined pool of prompts related to that metric. These metric-specific prompts are stacked in dedicated sections of the next prompt, organized by evaluation type (e.g., ``\texttt{LENGTH DIVERSITY GUIDELINES}''). Each section accumulates up to three active instructions per metric. When more than three prompt instructions are needed for a given metric, the oldest ones are replaced in a round-robin fashion to maintain a sliding window of guidance. This design avoids unbounded prompt growth, which could lead to instruction bloat and increase the risk of LLM prompt saturation—where the model begins to ignore or under-prioritize critical directives due to excessive or competing context. Limiting the number of instructions helps ensure that each metric receives focused, interpretable guidance while maintaining the overall prompt's effectiveness and coherence.

This process creates a feedback loop that fine-tunes the model's behavior over time: the model is repeatedly exposed to evolving instructions emphasizing failed dimensions, leading to improvements in review diversity, quality, and realism. The pipeline retains all generated outputs and never discards batches outright. Instead, it prioritizes iterative improvement while accumulating high-quality samples.

\section{Results and Discussion}
In order to validate whether the proposed metrics can effectively reflect the diversity and privacy of user reviews, we first apply the proposed metrics on a large-scale real user dataset. This real user dataset includes 2.5 million Amazon product reviews. We mainly focus on three columns of data, including user ID, rating value, and review text. The results validate that the proposed metrics can effectively reflect the diversity and privacy risks of the real user data. Due to page limit, the detailed results are included in Appendix C.

\begin{figure}[t]
    \centering
\includegraphics[width=1\linewidth]{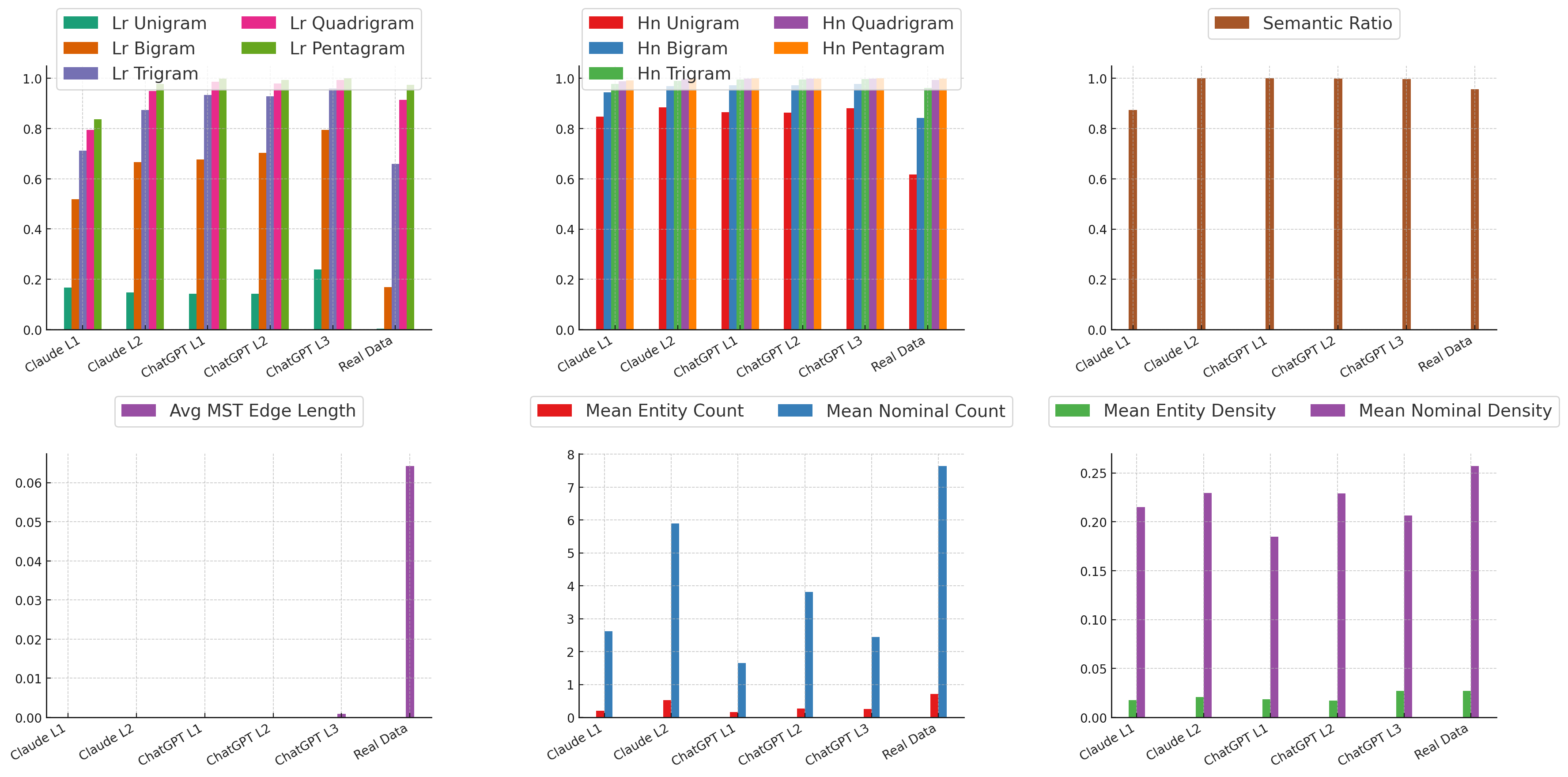}
    \caption{Visualization of results for every evaluation metric. See Appendix D for full detailed data table.}
    \label{fig:results_chart}
\end{figure}

We then applied the same metrics to synthetic datasets generated by the state-of-the-art LLMs, including GPT‑4o and Claude 3.7 Sonnet, to assess these models' generation capabilities in terms of review diversity and privacy. Three different levels of prompts are designed to generate multiple synthetic datasets for evaluation and comparison. 
\begin{itemize}
    \item \textbf{Prompt Level 1:} Basic prompt with minimal constraints.
    \item \textbf{Prompt Level 2:} Manually improved prompt with specific constraints.
    \item \textbf{Prompt Level 3:} Adaptive prompt iteratively optimized based on evaluation feedback.
\end{itemize}
The basic prompt is to generate a batch of synthetic product reviews with minimal constraints. Example reviews were provided to guide tone and sentiment distribution, while prohibiting direct copying to ensure originality and privacy. This initial prompt served as a baseline for evaluating text diversity and privacy sensitivity in the generated data. We then introduced a level 2 prompt with domain-specific focus (fashion and clothing) and stricter constraints to enhance realism. It emphasized lexical and semantic diversity, realistic sentiment patterns, and distinct content across reviews. The synthetic data from this prompt were evaluated using the same criteria as before. Level 3 prompt is the proposed adaptive prompt scheme, as discussed in Section \ref{sec:ada_prompt}, which adapts its instructions based on multi-metric feedback. The results are shown in Figure \ref{fig:results_chart}.

\paragraph{Privacy Analysis}
The synthetic data generated using the basic prompt tended to be shorter and more fragmented, with minimal word use and limited expressive detail. As a result, these reviews rarely contained sensitive terms or identifiable patterns of concern. In contrast, the level 2 prompt—which emphasized lexical and semantic diversity, realistic sentiment distributions, and distinct content—produced longer and more expressive reviews. This led to a noticeable increase in the use of pronouns and the emergence of sensitive terms, particularly those referencing family relationships (e.g., “my daughter”), body measurements (e.g., height, weight), and clothing sizes. For example, the original review “Purchased this bowtie for my daughter” was transformed into “Bought this beanie for my daughter and she hasn’t taken it off since”—a richer, more personal statement that introduces kinship and emotional connection. These findings suggest that while stricter generation constraints improve realism, they may also increase the likelihood of privacy-relevant content appearing in synthetic reviews.

The adaptive prompt produced the most realistic synthetic reviews to date, with a natural mix of short, minimal-word entries and longer, context-rich narrative reviews. However, this gain in realism came with increased privacy risks. We observed a notable rise in sensitive content, especially terms related to family and kinship, suggesting that as synthetic data becomes more lifelike, the likelihood of unintentionally generating privacy-relevant language also increases. Additionally, embedding-level patterns further highlighted potential privacy concerns. When applying a global z-score threshold of -1.0, users below this threshold showed significantly higher cosine distances to their nearest neighbors. This suggests that these users possess more distinct writing styles compared to the broader population, increasing the risk of re-identification or exposure of memorized language patterns.

\paragraph{Diversity Analysis} 

Upon analyzing the diversity evaluations, we first look at the generation from Claude Level 1 and 2. There was a substantial increase in the lexical uniqueness ratio with higher N-grams for prompt Level 2. These results suggest that the Level 2 prompt for Claude effectively produced more unique and diverse phrases at higher N-gram levels with a more detailed prompt. The semantic diversity metrics reveal a trade-off at the level 2 generation. The Average MST Edge Length significantly decreased, implying closer semantic relationships among reviews. However, Claude achieved a perfect Semantic Ratio of 1.0 at level 2, meaning there were no repeated reviews. Although the diversity metrics do not evaluate the length of words in each review, the level 2 prompt significantly increased the length of the reviews for both Claude and ChatGPT. The difference can be seen in these two example reviews from Claude level 1 and level 2 generation respectively, "This product is complete garbage. Broke within a day.", "These jeans exceeded my expectations. Perfect fit around the waist and the length is just right for my height. The material feels durable yet comfortable for all-day wear.". So while the semantic relationships between the reviews are closer together in the level 2 generation, the improvised prompt increased the length and distinctiveness of the reviews. It is also worth noting that the D$_{sen}$ score for sentiment diversity had no significant change.

When analyzing ChatGPT’s generated data, it demonstrated remarkable consistency in lexical diversity metrics between the two levels. The same semantic trend seen in Claude’s data was seen with ChatGPT, where the level 2 Average MST Edge Length decreased. However unlike Claude, the Semantic Ratio remained very similar. Again, a trend can be observed with ChatGPT’s level 2 data where the length of the reviews were noticeably longer than the level 1 generated data. A similar tradeoff is seen in which the reviews themselves became longer, but the semantic relationship became closer and less diverse. However, the lexical and sentiment diversity were able to remain very similar and show that a more descriptive prompt can maintain these two types of diversity.

The most notable improvements with the Level 3 prompt technique is the Average MST Edge Length (0.000914) and the N-gram L$_r$ scores. The average edge length greatly surpasses any of the previous averages from the other synthetic generation levels. This significant improvement along with the Semantic Ratio score of 0.9980 shows that this technique maintains the semantic uniqueness with almost every review being distinct, while the reviews are semantically farther away than the previous techniques. Also, the L$_r$ and H$_n$ scores for each of the N-gram levels are higher than the previous generation techniques. This increased L$_r$ score reveals a higher variability at the lexical level and less repeated words across this generated dataset. This score paired with the H$_n$ value shows the distribution is also just as evenly spread out as the previous generations, if not more. The adaptive prompt has shown significant increases in both semantic and lexical diversity, while maintaining individual review uniqueness.

\vspace{-4mm}

\section{Conclusion and Future Work}

This article presents a comprehensive synthetic customer review data generation system that produces unlimited, high-quality, and privacy-preserving review data. This data can be used to fine-tune third-party large language models (LLMs), enhancing our understanding of eBay customers and boosting customer satisfaction. Additionally, this approach can be broadly applied to anonymize and balance any free-form textual data, providing a valuable resource for pre-training any foundational LLM model.

While our current approach uses hand-crafted evaluation metrics and a static list of prompt refinements, future work could explore learning-based techniques for more efficient and intelligent prompt adaptation. One promising direction is to frame the prompt optimization problem as a reinforcement learning task, where the LLM acts as a black-box environment and prompt updates are actions selected to maximize reward (e.g., batch quality scores). 

Another extension involves incorporating additional evaluation metrics such as diversity in named entities, sentiment consistency across review subsets, or alignment with real-world data distributions. Finally, replacing manually curated prompt pools with a generative or learned mechanism could offer greater scalability and adaptability across domains.


\bibliographystyle{unsrt}  
\bibliography{references}  

\newpage
\appendixpage
\section*{Appendix A: Preprocessing Workflow}\label{sec:embedding}

To quantitatively assess the diversity and privacy of synthetically generated review data, this research implements a structured preprocessing and vectorization workflow, as shown in Figure \ref{fig:preprocessing_workflow}. The workflow begins with textual preprocessing, followed by word embedding, and culminates in a sentence-level embedding for each review. The initial preprocessing step involves removing stopwords such as "the", "is", and "at", which frequently appear in text but provide little meaningful semantic information. Stopword removal significantly reduces noise within the data, emphasizing the presence of meaningful, sentiment-rich words with clearer semantic representations. By eliminating these common but semantically insignificant words, the subsequent embedding process effectively captures the true content and emotional nuance of each review, improving the accuracy of downstream evaluation.

Following preprocessing, reviews are converted into numerical representations using the Word2Vec embedding model. This method was specifically selected due to its capacity to capture nuanced semantic relationships between words based on contextual co-occurrences within the review corpus itself. Unlike generalized pre-trained embedding models, which might introduce irrelevant external semantic biases, training Word2Vec specifically on the dataset under study ensures the embeddings precisely reflect the linguistic characteristics unique to the dataset. The sentences parameter explicitly provides the preprocessed tokens from the review dataset, meaning the Word2Vec model learns semantic relationships solely based on the vocabulary and word patterns present in these reviews. A key parameter of the Word2Vec model is the window size (denoted as $W_s$), which determines the number of neighboring words on either side that contribute to each word's embedding. In this work, we set $W_s=5$ to capture sufficient contextual information to establish meaningful semantic relationships, balancing the depth of semantic context against computational resources and the risk of capturing overly distant, potentially irrelevant context. Another key parameter is the dimensionality of the embedded vector of each word. In this study, it is set as 100 to lower computational complexity while maintaining sufficient representational complexity to capture meaningful semantic distinctions.

To represent each entire review quantitatively, a sentence-level embedding is computed by averaging the word vectors contained within each review. This averaging method was chosen because it provides a balanced and representative embedding that reduces the risk of distortion from individual outlier words or disproportionately influential terms. This ensures that sentence embeddings capture the overall semantic essence of the review, enabling accurate and interpretable comparisons between reviews in future steps. These comparisons are central to evaluating diversity and detecting redundancy within synthetic datasets, as they quantify the semantic similarity or dissimilarity among reviews. After the preprocessing, the embeddings will be fed into the diversity and privacy evaluation modules for further analysis.

\section*{Appendix B: Equations for proposed metrics}

\textbf{Lexical Uniqueness Ratio}: For each N-gram level, this ratio quantifies the lexical diversity by calculating the proportion of unique N-grams (distinct word combinations) relative to the total number of N-grams identified. A higher lexical uniqueness ratio indicates greater lexical richness and lower repetitiveness, critical for evaluating the variability and originality of synthetic text.

Mathematically, this is represented as:
    \begin{equation}
    L_{r} = \frac{U}{T}
    \end{equation}
where \(U\) is the number of unique N-grams, and \(T\) is the total count of N-grams at that specific N-gram level.

\textbf{Normalized Lexical Entropy}: Entropy evaluates the randomness or unpredictability within the distribution of N-grams at each N-gram level. The entropy H for a given N-gram level is calculated using the following formula:
    \begin{equation}
    H = -\sum_{j=1}^{U} p_j \log_2(p_j)
    \end{equation}
    \begin{equation}
    p_j = \frac{f_j}{T}
    \end{equation}
    where \(p_j\) represents the relative frequency (probability) of the \(j^{th}\) unique N-gram. \(p_j\) is computed with \(f_j\) being the frequency of the \(j^{th}\) N-gram, and \(T\) the total count of all N-grams for that respective N-gram level. For entropy H, U denotes the total number of unique N-grams for that N-gram level. 

To ensure comparability across different N-gram levels, entropy is normalized using the following formula:
    \begin{equation}
         H_{\text{n}} = \frac{H}{\log_2(U)}
    \end{equation}
    Normalized lexical entropy values range between 0 and 1, with higher values signifying more evenly distributed and diverse linguistic patterns.


\textbf{Named Entity Extraction}      
To extract named entities, we utilize the \texttt{en\_core\_web\_sm} model from the spaCy Natural Language Processing (NLP) library \cite{spacy2020}, which identifies entities such as persons, locations, organizations, and dates. The model processes each review and outputs all named entities along with their types. For each review, we extract two pieces of information: (1) the total count of named entities (i.e., $e_i$ for review $i$), and (2) the entity density (i.e., $\rho_i$ for review $i$), defined as the ratio of the number of named entities to the total number of tokens excluding whitespace and punctuations (i.e., $T_i$ for review $i$), as shown in Equation (\ref{eq:entity_density}).

\begin{equation}
\rho_i = \frac{e_i}{T_i} \label{eq:entity_density}
\end{equation}

To evaluate entity exposure across the entire dataset, we compute the following statistics:

\begin{itemize}
    \item Mean entity count: \( \mu_e = \frac{1}{N} \sum_{i=1}^{N} e_i \)
    \item Mean entity density: \( \mu_\rho = \frac{1}{N} \sum_{i=1}^{N} \rho_i \)
    \item Max entity count: \( \max_e = \max_{1 \leq i \leq N} e_i \)
    \item Max entity density: \( \max_\rho = \max_{1 \leq i \leq N} \rho_i \)
\end{itemize}

These statistics provide insight into the overall level of real-world referencing, as well as the most extreme cases that may present higher privacy risks.

\textbf{Nominal Mention Extraction}      
In addition to named entities, we extract nominal mentions, which are tokens that refer to specific roles or participants in the sentence. These include:
\begin{itemize}
    \item \textbf{Subjects} (e.g., ``I'', ``my daughter'')
    \item \textbf{Objects} (e.g., ``my son'', ``gift for him'')
    \item \textbf{Proper nouns} (e.g., ``Palo Alto'', ``Stanford'')
    \item \textbf{Pronouns} (e.g., ``she'', ``they'')
\end{itemize}
We identify these parts of speech and syntactic roles using spaCy’s dependency parser and part-of-speech tagger. For each review, we extract two pieces of information: (1) the total count of unique nominal tokens (i.e., $n_i$ for review $i$), and (2) the nominal density (i.e., $\delta_i$ for review $i$), defined as the ratio of unique nominal tokens to the total number of tokens excluding whitespace (i.e., $T_i$ for review $i$), as shown in Equation (\ref{eq:nominal_density}). These features provide insight into how heavily the text leans on participant references, which can indirectly reveal user intent, relationships, or demographic information.

\begin{equation}
\delta_i = \frac{n_i}{T_i} \label{eq:nominal_density}
\end{equation}

Similarly, for the overall evaluation, we compute the following four statistics:

\begin{itemize}
    \item Mean nominal count: \( \mu_n = \frac{1}{N} \sum_{i=1}^{N} n_i \)
    \item Mean nominal density: \( \mu_\delta = \frac{1}{N} \sum_{i=1}^{N} \delta_i \)
    \item Max nominal count: \( \max_n = \max_{1 \leq i \leq N} n_i \)
    \item Max nominal density: \( \max_\delta = \max_{1 \leq i \leq N} \delta_i \)
\end{itemize}

\textbf{Average Cosine Similarity for Each User}

We compute the average cosine similarity between each pair of embedding vectors at the user level. Let \( N \) denote the total number of users, and the average cosine similarity \( S_i \) for user \( i \) is calculated as:

\begin{equation}
S_i = \frac{1}{N - 1} \sum_{j = 1}^{N} \cos(u_i, u_j)
\end{equation}
\newpage
\section*{Appendix C: Metric Validation Results on Real User Data} \label{sec:realuser_results}

\subsection*{1. Diversity Evaluations}

\paragraph{Sentiment Diversity Analysis}

\begin{figure}[h]
\centering

\begin{subfigure}[c]{0.46\linewidth}
    \includegraphics[width=\linewidth]{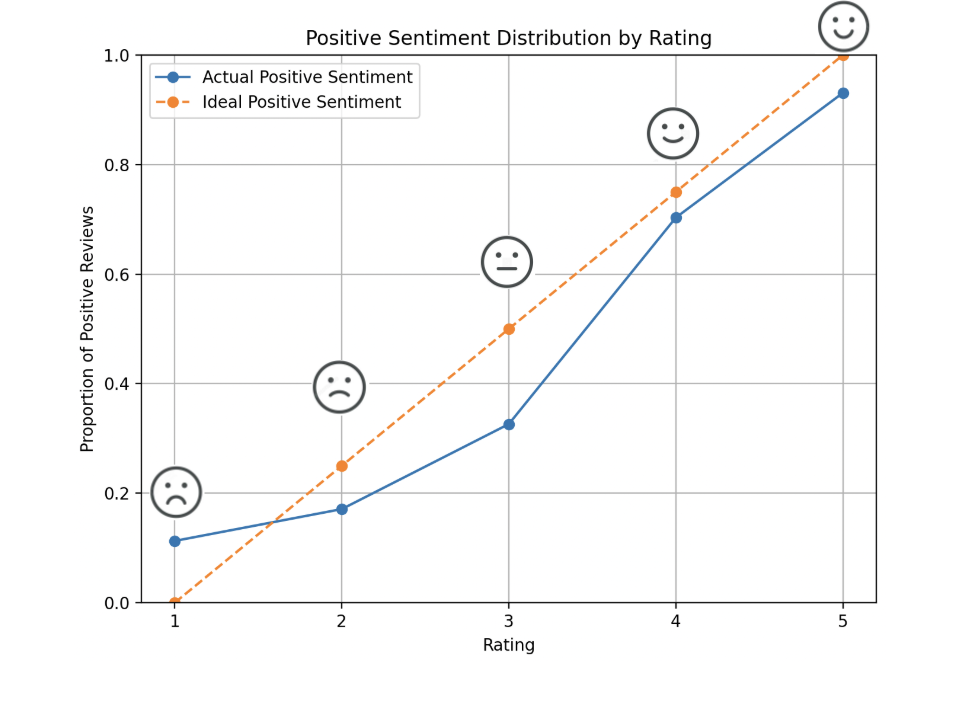}
    \caption{Positive sentiment distribution of the real user data (in blue) compared to the ideal linear distribution (in orange).} 
    \label{fig:sentiment_curve}
\end{subfigure}
\hfill
\begin{subfigure}[c]{0.49\linewidth}
    \centering
    \renewcommand{\arraystretch}{1.5}
    \begin{tabular}{|c|c|c|}
        \hline
        \textbf{Rating} & \textbf{Negative} & \textbf{Positive} \\ \hline
        1               & 88.72\%           & 11.28\%           \\ \hline
        2               & 82.91\%           & 17.09\%           \\ \hline
        3               & 67.38\%           & 32.62\%           \\ \hline
        4               & 29.68\%           & 70.32\%           \\ \hline
        5               & 6.91\%           & 93.09\%           \\ \hline
        \multicolumn{3}{|c|}{\textbf{D$_{sen}$ score: 0.9036}} \\ \hline
    \end{tabular}
    \caption{Sentiment distribution for real user data values and calculated D$_{sen}$ Score}
    \label{fig:sentiment_table}
\end{subfigure}

\caption{Graphical and tabular representation of real user data for sentiment diversity evaluation.}
\label{fig:sentiment}
\end{figure}

Upon analyzing the sentiment diversity of real user reviews, we observed a trend that aligns with the ideal sentiment distribution line established as a benchmark. The provided graph and accompanying table in Figure \ref{fig:sentiment} clearly demonstrate this alignment, showing that the proportion of positive sentiment reviews consistently increases with higher star ratings.

The calculated sentiment diversity score D$_{sen}$ from Equation \ref{eq:MAE_sum} is approximately 0.9036, indicating that the real user reviews strongly adhere to the ideal sentiment distribution. This high score underscores the fact that actual human-generated reviews generally reflect the natural trend where higher star ratings correlate strongly with more positive sentiments. However, the sentiment alignment is not exactly 1.0, showing some discrepancies in real-world data. Such discrepancies arise due to the nuanced nature of human reviews, where subjective expressions of sentiment do not always strictly align with rating values. Notably, the most pronounced deviation occurs at the 3-star rating, where the positive sentiment is 32.62\%, significantly lower than the benchmark’s neutral value of 50\%. This deviation is understandable, given that 3-star ratings often represent uncertainty, capturing mixed or even slightly negative sentiment rather than pure neutrality. For example, one real human review with a 3-star rating stated “Fell apart after a couple washes.”, which captures a negative sentiment. Logically, if a reviewer is critiquing the product and leaving a review of 3 instead of a higher value of 4 or 5, there can often be negative sentiment found in the text review. The slight deviations at other rating levels, such as the 1-star category, which recorded an 11.28\% positivity rate further demonstrate the complexity of human sentiment. Even negative ratings sometimes contain positive aspects, as seen in this example which had a rating of 1, “Nice job..ordered the name `Samantha' got the name `Ava'..”, where the reviewer wrote a sarcastic statement that was interpreted with positive sentiment. 

\paragraph{Word-level Lexical Diversity Analysis}

Analyzing the lexical diversity of the real user reviews dataset reveals significant insights across different N-gram levels. Starting with the unigrams (single words), the lexical uniqueness ratio is notably low at 0.0042. This suggests a high repetition rate, with many words frequently recurring across the reviews. This result can be expected, as individual words naturally recur more frequently within a large corpus, particularly common adjectives, nouns, and descriptive terms related to the product being reviewed. As seen in Table \ref{tab:sensitive_examples}, some of the example reviews repeat certain N-grams multiple times in the same review such as the word “Hawaii” or “plantar fasciitis”. Correspondingly, the normalized lexical entropy for the unigrams is lower at 0.6176, indicating uneven distribution and imbalance in word usage. This imbalance arises from common words significantly overshadowing less frequently used words, causing a skew in the distribution.

As we move to higher-order N-grams, the lexical uniqueness ratio increases substantially. For instance, the bigrams exhibit a ratio of 0.1696, significantly higher than that of unigrams, reflecting reduced repetition and increased distinctiveness. The trend continues progressively, with the trigrams showing a substantial jump in the lexical uniqueness ratio to 0.6588, followed by further increases for quadrigrams (0.9133) and pentagrams (0.9743). The increasing ratios demonstrate that higher level N-grams are inherently more unique and repeated far less frequently across the dataset.

Similarly, the normalized lexical entropy values increase with the N-gram size, approaching near-perfect scores as the N-gram order grows. For example, the normalized lexical entropy rises from 0.8419 for bigrams to as high as 0.9987 for pentagrams. These high values indicate a more even distribution among higher-order N-grams, highlighting the diverse usage of phrases and expressions. This trend emerges because longer phrases naturally possess greater uniqueness and less repetition compared to individual words or shorter combinations.

\paragraph{Review-level Semantic Diversity Analysis}

In assessing the review-level semantic diversity of the real user review dataset, which comprises approximately 2.5 million reviews, a full computation of the Minimum Spanning Tree (MST) proved computationally challenging due to the scale involved. Specifically, generating and processing a complete pairwise cosine distance matrix of size 2.5 million by 2.5 million was unfeasible. Therefore, we adopted a simplified yet effective approach by constructing an approximate MST using the nearest neighbor method with k = 30 neighbors. Although this method does not yield an exact MST, it provides a valid and computationally practical approximation of the semantic diversity inherent in the large dataset.

\begin{figure}[t]
\centering

\begin{subfigure}[c]{0.48\linewidth}
   \centering
    \renewcommand{\arraystretch}{1.5}
    \begin{tabular}{|c|c|c|}
        \hline
        \textbf{N-gram Level} & \textbf{Lexical Uniqueness} & \textbf{Normalized} \\ 
        &\textbf{Ratio}& \textbf{Lexical Entropy}\\\hline
        Unigram               & 0.0042           & 0.6176           \\ \hline
        Bigram                & 0.1696           & 0.8419           \\ \hline
        Trigram               & 0.6588           & 0.9619           \\ \hline
        Quadrigram            & 0.9133           & 0.9937          \\ \hline
        Pentagram             & 0.9743           & 0.9987           \\ \hline
    \end{tabular}
    \caption{Word-level lexical richness evaluation on real user data}
    \label{fig:lexical_evaluation_table}
\end{subfigure}
\hfill
\begin{subfigure}[c]{0.45\linewidth}
    \centering
    \renewcommand{\arraystretch}{1.5}
    \begin{tabular}{|c|c|}
        \hline
        \textbf{Evaluation Metric} & \textbf{Score} \\ \hline
        Average MST Edge Length    & 0.0642          \\ \hline
        Semantic Ratio             & 0.9672           \\ \hline
    \end{tabular}
    \caption{Review-level semantic diversity evaluation summary of real user reviews.}
    \label{fig:semantic_evaluation_table}
\end{subfigure}

\caption{Lexical \& Semantic diversity Evaluations on Real User Reviews.}
\label{fig:semantic}
\end{figure}


The Average MST Edge Length measures the average semantic distance between reviews directly connected within the MST, considering only distinct nonzero edges. In this analysis, an average edge length of 0.064 as seen in Figure \ref{fig:semantic_evaluation_table}, indicates that on average, each review (node) in the MST connects to other reviews that are relatively semantically close. While this value might initially appear low, it is important to recognize that the MST inherently represents the minimal set of edges required to connect all nodes. Consequently, MST edge lengths are often shorter because the construction prioritizes minimal semantic distances, capturing fundamental semantic relationships rather than maximal differences. That being said, in natural human reviews, there are often overlapping or similar reviews in a given dataset. 

However, interpreting the average edge length alone could be misleading without additional context. For example, a dataset with many repeated reviews could yield a deceptively high average edge length because only the distinct nodes are calculated for the average. Therefore, the introduction of the Semantic Ratio (0.96 in this dataset) is crucial. This ratio quantifies the proportion of distinct semantic embeddings relative to the total reviews, thus validating the representativeness of the average edge length. A high semantic ratio (such as 0.96) confirms that most reviews are distinct, enhancing confidence that the calculated average edge length accurately reflects genuine semantic diversity rather than artifacts from numerous repetitions.


\subsection*{2. Analysis of Named Entities and Nominal Mentions}
Upon inspecting the review dataset, we observe that the types of sensitive information surfaced by named entity and nominal mention extraction tend to cluster into five common categories: (1) personal names, (2) kinship or family member references (e.g., “my daughter,” “grandson”), (3) physical attributes and health descriptors, such as body height, weight, or medical conditions, (4) clothing sizes and fit-related details, and (5) geographic or location-specific mentions. These categories reflect real-world identifiers that may increase privacy risk when preserved or reproduced by downstream language models. As anticipated, reviews with high entity counts and nominal counts are generally longer and provide rich contextual detail. These reviews often include dense references to personal experiences, making them more likely to contain identifiable content.

\textbf{Entity-dense reviews.} In high entity count cases, we observe a consistent pattern: sensitive terms frequently belong to categories (1) personal names, (4) clothing sizes, and (5) geographic locations, with partial coverage of (2) kinship references and (3) physical attributes. For instance, age references like “thirteen-year-old” (typically labeled as \texttt{DATE}) and body metrics such as “160 lbs” or “5 ft 8” (often labeled as \texttt{CARDINAL}) are commonly extracted. These phrases, while not direct identifiers on their own, can contribute to re-identification risk when combined with other contextual details, especially in longer reviews where multiple such cues co-occur.

\textbf{Pronoun-heavy reviews.} High nominal count reviews, on the other hand, are typically characterized by the extensive use of pronouns and grammatical references to the self or others (e.g., “I,” “my,” “he,” “they”). While individual pronouns may seem innocuous, their repeated use often signals narrative structure and involvement of specific participants, which can implicitly reveal roles, relationships, and user perspective. When combined with the surrounding textual context, these references may enable associative inference about the reviewer’s identity, role, or circumstances, particularly when the review references specific actions or interactions with named individuals.

Reviews with high entity density or high nominal density tend to be extremely short, often containing fewer than ten words. This aligns with our expectations, as approximately 30\% of all reviews in the dataset fall below this length threshold. However, in these short and fragmented reviews, we observed a significant decline in the performance of spaCy's named entity recognition. Among 1{,}000 high-entity-density reviews sampled for manual inspection, only 10–20\% contained genuinely sensitive terms; the remainder were primarily false positives triggered by capitalization. For instance, in phrases such as “Cheaply made”, the word “Cheaply” was incorrectly tagged as an \texttt{ORG} (organization). These errors stem from spaCy's reliance on syntactic and semantic context: its entity recognizer leverages contextual cues from neighboring tokens, sentence structure, and linguistic patterns, which are often absent or degraded in short, informal reviews. Without sufficient context, the model tends to default to surface features such as capitalization, resulting in unreliable entity predictions. In contrast, reviews with high nominal density showed relatively fewer issues. Because pronouns are generally less dependent on extended context, they were extracted with higher reliability. Nonetheless, we observed occasional misclassifications of capitalized adjectives or interjections as proper nouns—for example, in “Super Cute!”, both “Super” and “Cute” were incorrectly labeled as \texttt{PROPN}. These limitations highlight the challenge of applying general-purpose NLP models to informal, user-generated text with minimal structure.

The examples in Table~\ref{tab:sensitive_examples} were selected to represent four distinct scenarios of privacy-relevant content: high entity count, high nominal count, high entity density, and high nominal density. Collectively, these examples capture a range of privacy-relevant patterns—from long, context-rich reviews containing detailed personal references, to short but dense reviews in which even a few words can reveal relationships or sensitive traits. These include references to educational background (e.g., “University of Hawaii”, “Bachelor’s degree”), geographic location (e.g., “Hawaii”), clothing size (e.g., “XL”, “large/extra-large”), physical attributes such as height and weight (e.g., “5'8”, “110”), medical conditions (e.g., “plantar fasciitis”), family relationships (e.g., “granddaughter”, “boyfriend”), and purchasing context (e.g., “9 to 9.5 shoe”, “Amazon”, “\$12.99 to \$16.99”).

\subsection*{3. User-Level Stylistic Profiling and Outlier Analysis}
\paragraph{Distribution of Average Cosine Similarities}

\begin{figure}[t]
\centering
\includegraphics[width=0.7\linewidth]{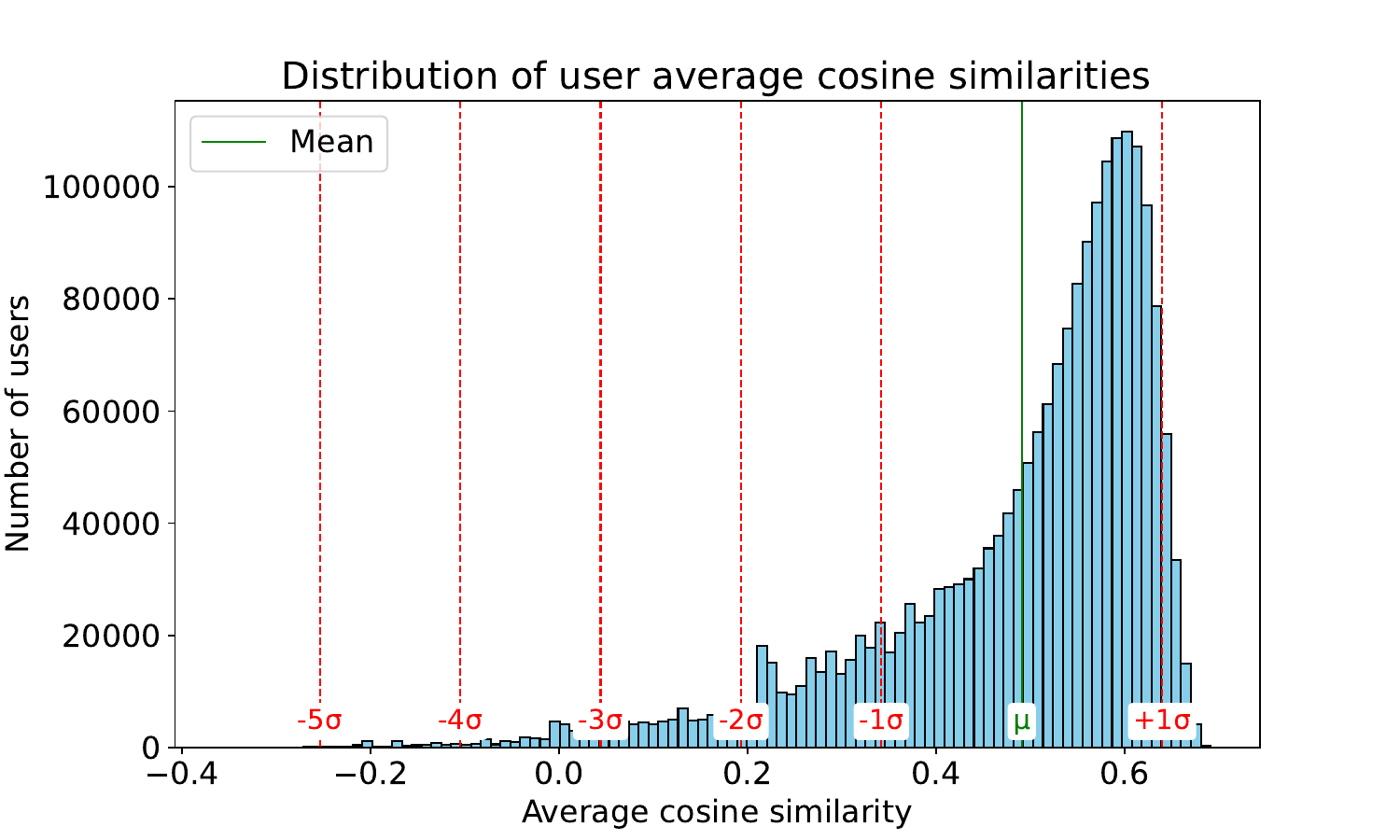}
\caption{Distribution of user average cosine similarities, with vertical lines marking the mean (\(\mu\)) and standard deviations (\(\pm1\sigma\), \(\pm2\sigma\), etc.). Users with low similarity scores (e.g., beyond \(-2\sigma\)) are potential stylistic outliers.}
\label{fig:similarity_histogram}
\end{figure}

As a first step in identifying stylistic outliers, we examined the distribution of average cosine similarity scores between each user and all others in the dataset. Figure~\ref{fig:similarity_histogram} shows this distribution, where each value reflects how similar a user's writing style is to the rest of the population. The distribution is left-skewed, with most users clustered around higher similarity values and a long tail extending toward lower similarities. This tail includes users who are stylistically distant from others and are candidates for further inspection.

\begin{figure}[t]
\centering
\begin{subfigure}[t]{0.49\linewidth}
    \includegraphics[width=\linewidth]{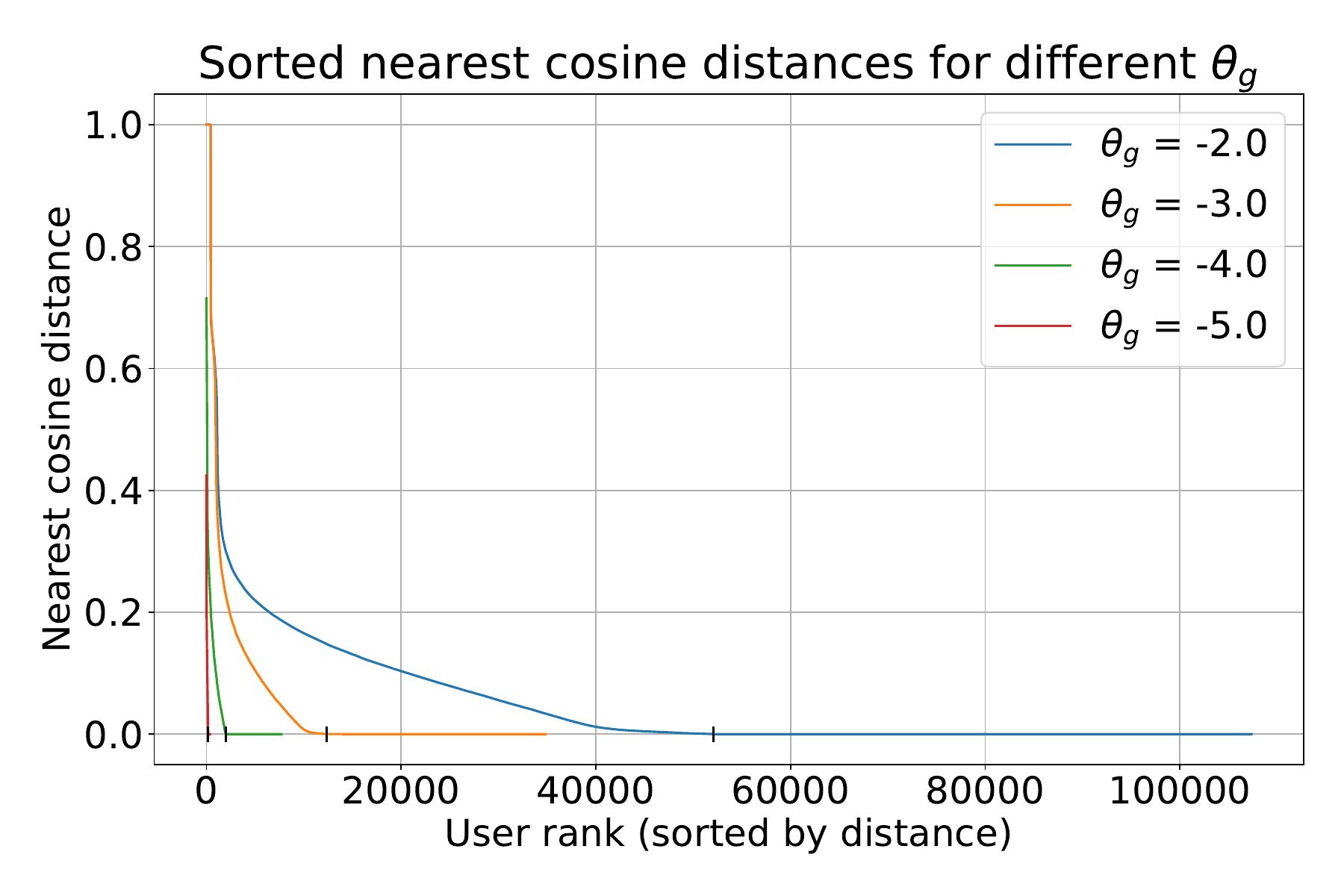}
    \caption{Nearest neighbor distance curves for a range of global thresholds from $\theta_g \in [-2, -5] $, illustrating the effect of increasingly strict global filtering.}
    \label{fig:local_distance_extended}
\end{subfigure}
\hfill
\begin{subfigure}[t]{0.49\linewidth}
    \includegraphics[width=\linewidth]{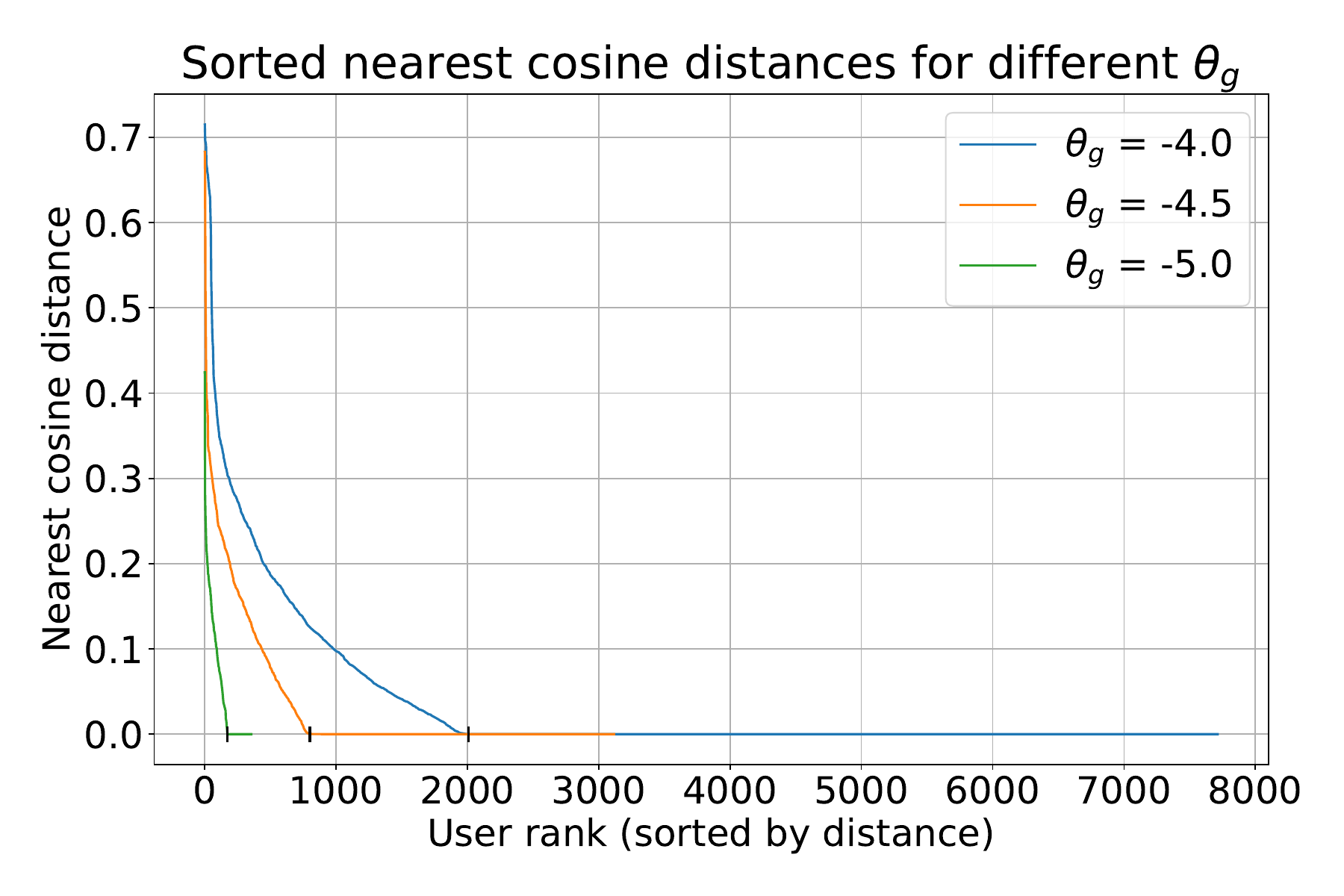}
    \caption{The curves of the low thresholds $\theta_g \in \{-4.0, -4.5, -5.0\}$, highlighting the separation between locally isolated users and those near stylistic clusters.}
    \label{fig:local_distance_zoom}
\end{subfigure}
\caption{Nearest neighbor distance curves for globally distant users under varying global thresholds.}
\label{fig:local_distance}
\end{figure}

\paragraph{Nearest Neighbor Distance Analysis}
To assess how isolated the globally distant users are, we analyzed their nearest neighbor distances in the embedding space. Figure~\ref{fig:local_distance_extended} and Figure~\ref{fig:local_distance_zoom} show sorted nearest neighbor distances for users under progressively stricter global thresholds (i.e., $\theta_g \in [-2, -5] $). The sorted nearest neighbor distance curves exhibit a two-part structure for each global threshold. The first part consists of a steep drop from relatively high cosine distances (often close to 1.0) down to near zero, followed by a flat tail where the remaining users have very small distances to their nearest neighbors. As the global threshold becomes more extreme (e.g., $\theta_g \in \{-4.0, -4.5, -5.0\}$), this two-part shape becomes increasingly distinct. In our analysis, we consider users in the steep initial segment of each curve to be the final stylistic outliers, as they are both globally and locally distant—far from the general population and not closely aligned with any nearby stylistic cluster. To define the boundary between the steep and flat regions of each curve, we used the local distance threshold $\theta_l =10^{-4}$ as the flattening cutoff. This threshold is adjustable and can be tuned depending on the sensitivity required for identifying stylistic outliers.

\newpage
\section*{Appendix D: Results of LLM-generated Synthetic Data}

\subsection*{Level 1 Prompt}

You are an advanced AI model tasked with generating synthetic product reviews. Below are several examples of product reviews, each with multiple attributes. Your goal is to generate new, synthetic product reviews that follow the same structure and maintain a similar distribution in terms of ratings and content style. The synthetic reviews should not duplicate or leak any specific information from the provided examples. Instead, they should demonstrate creativity while adhering to the patterns and themes observed in the examples.

\paragraph{Example Reviews}\mbox{}\\
rating,review,user-id\newline
<Place example reviews here>

\paragraph{Instructions}
\begin{itemize}
    \item Generate $\{num\_reviews\}$ new product reviews.
    \item Each review should include the following attributes: rating, text, user-id.
    \item Ensure the synthetic reviews reflect the diversity and distribution of ratings (1.0 to 5.0) present in the example reviews.
    \item Do not copy any specific details or phrases from the example reviews.
    \item The synthetic reviews should be plausible and relevant to a variety of products.
    \item Maintain the style and tone observed in the example reviews.
\end{itemize}

Please generate the synthetic product reviews now.

\subsection*{Level 2 Prompt}

You are an advanced AI model tasked with generating synthetic product reviews. Below are several examples of fashion and clothing product reviews, each with multiple attributes. Your goal is to generate new, synthetic product reviews that follow the same structure and maintain a similar distribution in terms of ratings and content style. The synthetic reviews should not duplicate or leak any specific information from the provided examples. Instead, they should demonstrate creativity while adhering to the patterns and themes observed in the examples. 

\paragraph{Example Reviews}\mbox{}\\
rating,review,user-id\newline
<Place example reviews here>

\paragraph{Instructions}
\begin{itemize}
    \item Generate $\{num\_reviews\}$ new reviews about fashion and clothing products.
    \item Each review must include attributes: rating, review, user-id.
    \item Reflect realistic diversity and distribution of ratings (1.0 to 5.0).
    \item Avoid copying details or phrases from the example reviews.
    \item Generate plausible reviews relevant to various product items.
    \item Maintain realistic style, tone, and sentiment diversity.
\end{itemize}

\paragraph{Important Constraints}
\begin{itemize}
    \item Ensure reviews reflect a realistic and balanced distribution of sentiments for each rating, matching typical customer review patterns.
    \item Maintain high lexical diversity (unique phrasing, varied vocabulary). Use varied language, synonyms, and expressions. Avoid repeating words, phrases, or sentence structures.
    \item Ensure semantic distinctness (each review must describe unique experiences or product aspects). Ensure each review describes unique experiences, products, or aspects distinctly different from each other.
    \item Avoid generating reviews that appear unrealistic, generic, or disconnected from typical customer feedback.
\end{itemize}

Please generate the synthetic fashion and clothing product reviews now.

\begin{figure}[t]
    \centering
    \small  
    \renewcommand{\arraystretch}{1.3}
    \begin{adjustbox}{center, max width=\linewidth}
    \begin{tabular}{|c|c|c|c|c|c|c|}
        \hline
        \textbf{Evaluation Metrics} & \textbf{Claude L1} & \textbf{Claude L2} & \textbf{ChatGPT L1} & \textbf{ChatGPT L2} & \textbf{ChatGPT L3} & \textbf{Real Data} \\ \hline
        Unigram (L$_r$/H$_n$)       & 0.1672 / 0.8474 & 0.1480 / 0.8847 & 0.1419 / 0.8653 & 0.1422 / 0.8640 & 0.2386 / 0.8813 & 0.0042 / 0.6176  \\ \hline
        Bigram                      & 0.5184 / 0.9439 & 0.6657 / 0.9692 & 0.6768 / 0.9721 & 0.7025 / 0.9729 & 0.7942 / 0.9786 & 0.1696 / 0.8419  \\ \hline
        Trigram                     & 0.7114 / 0.9780 & 0.8739 / 0.9904 & 0.9335 / 0.9957 & 0.9274 / 0.9951 & 0.9592 / 0.9967 & 0.6588 / 0.9619  \\ \hline
        Quadrigram                  & 0.7939 / 0.9876 & 0.9488 / 0.9970 & 0.9857 / 0.9993 & 0.9798 / 0.9988 & 0.9929 / 0.9996 & 0.9133 / 0.9937  \\ \hline
        Pentagram                   & 0.8372 / 0.9915 & 0.9777 / 0.9988 & 0.9991 / 1.0000 & 0.9925 / 0.9996 & 0.9993 / 1.0000 & 0.9743 / 0.9987  \\ \hline
        Avg MST Edge Length         & 0.000043        & 0.000018        & 0.000078        & 0.000026        & 0.000914 & 0.0642            \\ \hline
        Semantic Ratio              & 0.8740          & 1.0000          & 1.0000          & 0.9990          & 0.9980 & 0.9572            \\ \hline
        D$_{sen}$ Score             & 0.8999          & 0.8938          & 0.9211          & 0.9234          & 0.9055 & 0.9036            \\ \hline
        Mean entity count           & 0.2040          & 0.5260          & 0.1636          & 0.2650          & 0.2619 & 0.7106            \\ \hline
        Mean nominal count          & 2.6180          & 5.8930          & 1.6534          & 3.8184          & 2.4506 & 7.6358            \\ \hline
        Mean entity density         & 0.0177          & 0.0206          & 0.0187          & 0.0172          & 0.0272 & 0.0269            \\ \hline
        Mean nominal density        & 0.2151          & 0.2295          & 0.1849          & 0.2291          & 0.2063 & 0.2569            \\ \hline
        1st percentile $d^{nn}_i$   & $1.485\!\times\!10^{-4}$ & $3.463\!\times\!10^{-5}$ & $2.887\!\times\!10^{-4}$ & $6.491\!\times\!10^{-5}$ & 0.0064 & 0.1847 \\ \hline
    \end{tabular}
    \end{adjustbox}
    \caption{Evaluation metrics comparing model generations and real data.}
    \label{fig:diversity_evaluations_table}
\end{figure}

\subsection*{}
\end{document}